\documentclass[journal]{IEEEtran}
\usepackage{graphicx}
\usepackage{subcaption}
\usepackage{multicol}
\usepackage{array}
\usepackage{multirow}

\ifCLASSINFOpdf
\else
\fi

\begin{document}
%
\title{Improving Facial Emotion Recognition Systems Using Gradient and Laplacian Images}
%
%
%

\author{Ram Krishna Pandey,~Member IEEE, Souvik Karmakar, A G Ramakrishnan,~Senior Member IEEE and N Saha 
\thanks{Ram Krishna Pandey, Souvik Karmakar and A G Ramakrishnan are with the Indian Institute
of Science, Bangalore, India. Email ids: \{ramp, souvikk, agr\}@iisc.ac.in}
\thanks{N Saha is with NIT Warangal, India. Email id: snabagata@nitw.ac.in }
}

%
%

\markboth{}%
{Shell \MakeLowercase{\textit{et al.}}: }
%



\maketitle

\begin{abstract}
In this work, we have proposed several enhancements to improve the performance of any facial emotion recognition (FER) system. We believe that the changes in the positions of the fiducial points and the intensities capture the crucial information regarding the emotion of a face image. We propose the use of the \textit{gradient} and the \textit{Laplacian} of the input image together with the original input into a convolutional neural network (CNN). \textit{These modifications help the network learn additional information from the gradient and Laplacian of the images. However, the plain CNN is not able to extract this information from the raw images.} We have performed a number of experiments on two well known datasets KDEF and FERplus. Our approach enhances the already high performance of state-of-the-art FER systems by 3 to 5\%.


\end{abstract}


%
\IEEEpeerreviewmaketitle

\section{Introduction}

Machine recognition of human emotions is an important, interesting and challenging artificial intelligence problem. Emotions play an important role in interactions/communication and help others in understanding human behaviour and intention. Human emotions can be recognized from voice \cite{SER}, body language, facial expression and electroencephalography \cite{asymmetry}. However, facial expression forms a simpler and more powerful way of recognizing emotions. Facial expression or emotion adds a lot of information or context to our verbal communication. Humans can recognize facial expressions with significant accuracy and in a short time. Excluding neutral, there are seven types of human emotions that are recognized universally: anger, disgust, fear, happiness, sadness, surprise and contempt. In certain situations, humans are known to express more than one emotion at a time. Developing intelligent systems for facial emotion recognition (FER) has wide application in various areas such as clinical practice, human-computer interaction, behavioural science, virtual reality, augmented reality, entertainment and advanced driver assistant systems. Traditional techniques for FER mainly consist of four successive steps: (i) pre-processing (ii) face and landmark detection (iii) feature extraction (iv) emotion classification. These approaches heavily depend on the algorithms used for face detection, landmark detection, the handcrafted features and the classifiers used. Recent developments in deep learning reduce the burden of handcrafting the features. Deep learning approaches perform well for all the above-mentioned tasks by learning an end-to-end mapping from the input data to the output classes. Out of all the learning based techniques, convolutional neural network (CNN) based techniques are preferred, where the extracted features are combined using a dense layer and the expressions are classified based on the output score of the soft-max layer. 

There is a tendency in researchers today to design deep neural networks (DNN's) as end to end systems, where every kind of processing is accomplished by the network, including the feature extraction, by learning from the data. Some people even opine that there is no need for any hard-coded feature extraction in any machine learning system. However, artificial neural networks and consequently, the deep neural networks have been designed in trying to simulate the biological neural network in the brain. Further, it is a well-known fact that there are many hard-coded feature extractors in the human, and even animal sensory systems, in addition to the natural neural network, that also learns from data (exposure and experience). For example, one might argue that it is quite possible for the visual neural pathway or cortex to extract the gray image from the colour image obtained by the cones in the retina. However, nature has chosen to have many more rods than cones to directly obtain the gray images also in parallel. Further, the Nobel prize winning work of Hubel and Wiezel \cite{hubel-wiesel} clearly showed the existence of orientation selective cells in the lateral geniculate nucleus and visual cortex of kitten. Also, different regions of the basilar membrane in the cochlea respond to different frequencies \cite{freqposfn} in both man and animals and this processing can be called as sub-band decomposition of the input audio signal. Thus, there are many examples of hard-coded feature extraction in the brain, and our work reported here, is inspired from this aspect of nature's processing. 
In this work, our main contributions are:
\begin{itemize}

	\item We have provided strategies (results listed in Tables~\ref{ourskdef},~\ref{vggoriginalwithoursresults}) and developed classifiers (results listed in Table~\ref{ourferparallel}) to improve the performance of any existing FER system.  
	
	\item We advocate the use of gradient and/or Laplacian to any FER system. This improves classifier recognition accuracy by a good margin (see Tables~\ref{ourskdef},~\ref{vggoriginalwithoursresults} and~\ref{ourferparallel}). 
	
	\item We have trained multiple models to validate the performance gain obtained due to the addition of gradient and Laplacian, on datasets KDEF~\cite{KDEF1998} and FERPlus~\cite{ferplus}. 
	
    \item The advantage of using the gradient and Laplacian of the input image is that the network extracts more features, which, together with the original features, capture changes in the image landmark points and their intensities. These changes are the important features for emotion recognition. Also, it obviates largely the need for extracting the fiducial or the landmark points.
		
	\item The plain CNN is not able to extract some information from the raw image and requires gradient and Laplacian for improved FER performance. We believe that our proposed approach can similarly improve the performance of any FER technique based on deep learning.

\end{itemize}
 
\section{Related Work}
Darwin suggested that human and animal facial emotions are evolutionary~\cite{darwin1998expression}. Motivated by Darwin's work, Ekman et al. \cite{ekman1969nonverbal}~\cite{ekman1970universal} found that the seven expressions, namely happiness, anger, fear, surprise, disgust, sadness and contempt remain the same across different cultures. Facial action coding system (FACS) is proposed in~\cite{friesen1978facial} to investigate the facial expressions and the corresponding emotions described by the activity of the atomic action units (cluster of facial muscles). Facial expression can be analyzed by mapping facial action units for each part of the face (eyes, nose, mouth corners) into codes.

\subsection{Traditional approaches}
Features are desired that possess maximal inter-class and minimal intra-class variabilities for each of the expressions. Traditional systems for facial emotion recognition are based on approaches that depend mainly on how the features are extracted from the facial expression. The features extracted can be categorized into (i) geometric features, (ii) appearance based features or (iii) their combination. In the work reported by Myunghoon et al~\cite{suk2014real}, facial features are extracted by active shape model, whereas Ghimire and Lee \cite{ghimire2013geometric} extract geometric features from the sequences of facial expression images and multi-class Ada-boost and SVM classifiers are used for classification. Global face region or regions containing different facial information are used to extract appearance-based features. Gabor wavelets, Haar features, local binary pattern~\cite{happy2012real} or its variant such as~\cite{zhao2007dynamic} are used to extract appearance-based features.
Ghimire and Lee \cite{ghimire2017facial} proposed a single frame classification of emotion using geometric as well as appearance based features and SVM classifier. In~\cite{2} features are extracted using pyramid histogram of gradients. Here, the facial edge contours are constructed using Canny edge detector. Histograms are calculated by dividing the edge maps into different pyramid resolution levels. The histogram vectors are concatenated to generate the final feature to be used for classification using SVM or AdaBoost classifier.
Boosted local binary patterns (LBP)~\cite{3} is another traditional image processing technique used for FER. A face is divided into small regions from which LBP histogram features are computed. For a given class, a template is generated from the histogram of facial features. Then a nearest neighbour classifier is used for classifying a given image.

\subsection{Deep learning based approaches}

\begin{figure}[ht!]
    \centering
\includegraphics[width=0.49\textwidth,height=0.11\textheight]{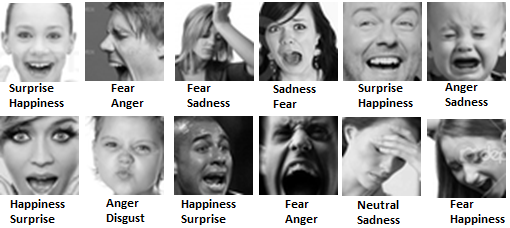}
\caption{Face image samples from FERplus dataset, with multiple emotion labels for each image~\cite{19}}
    \label{FER}
\end{figure}
The above techniques in the literature depend heavily on handcrafted features. However, deep learning algorithms have shown promising results in the recent years. CNN based models have shown significant performance gain in various computer vision and image processing tasks, such as image segmentation, de-noising, super-resolution, object recognition, face recognition, scene understanding and facial emotion recognition. Unlike the traditional techniques, deep learning based techniques learn (``end-to-end") to extract features from the data. For FER, the network generally uses four different kinds of layers, namely convolution, max-pool, dense layer and soft-max. Batch normalization with skip connection is also used to ease the training process. The features extracted have information about local spatial relation as well as global information. The max-pool layer helps in making the model robust to small geometrical distortion. The dense and soft-max layers help in assigning the class score.

Breuer et al.~\cite{breuer2017deep} demonstrate the capability of the CNN network trained on various FER datasets by visualizing the feature maps of the trained model, and their corresponding FACS action unit. Motivated by Xception architecture proposed in~\cite{chollet2016xception}, Arriaga et al. \cite{codepaper} proposed mini-Xception. Jung et al~\cite{jung2015joint} proposed two different deep network models for facial expression recognition. The first network extracts temporal appearance features, whereas the second extracts temporal geometric features and these networks are combined and fine tuned in the best possible way to obtain better accuracy from the model. Motivated by these two techniques, we have trained and obtained multiple models, the details of which are explained in section \ref{experiments}.

\begin{figure}
		\centering
		{   
	\includegraphics[width=0.15\textwidth,height=0.13\textheight]{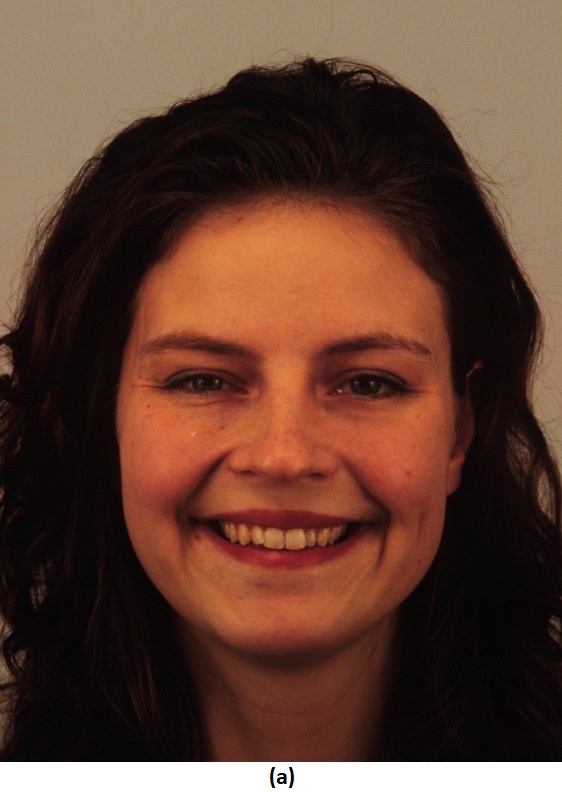}
	\includegraphics[width=0.15\textwidth,height=0.13\textheight]{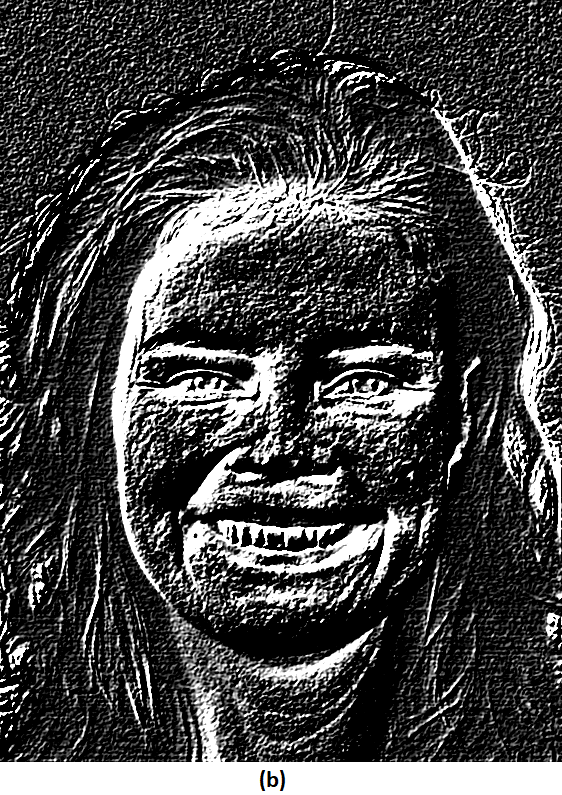}
	\includegraphics[width=0.15\textwidth,height=0.13\textheight]{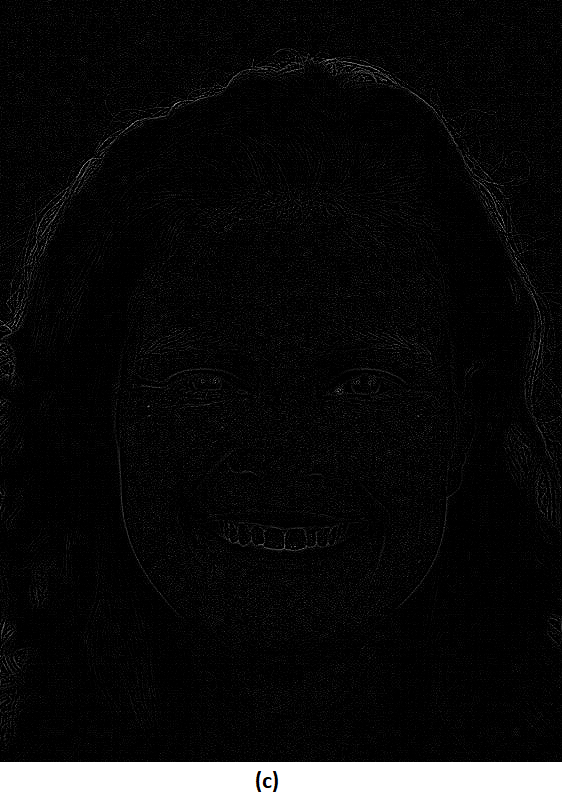}
			
	\caption{(a) A sample input image from the KDEF dataset. (b) Its derivative image obtained by the Sobel operator (Gradient). (c) Its second derivative obtained by the Laplacian operator. Zoom to see the details in the Laplacian image.}
	\label{result3}
		}
\vspace{-0.1 in}
\end{figure}

For the task of facial emotion recognition, the current state of the art model \cite{9} proposed a miniature version of VGG net, called VGG13. The network has 8.75 million parameters. The dataset used is the FERplus dataset, which has 8 classes, adding neutral to the existing seven classes. The reported test accuracy is $\approx84\%$. In 2014, G. Levi et al.~\cite{10} improved emotion recogition using CNN. They convert images to  local binary patterns. These patterns are mapped to a 3D metric space and used as an input to the existing CNN architectures, thus addressing the problem of appearance variation due to illumination. They trained the existing VGG network~\cite{5}, on CASIA Webface dataset, and then used transfer learning to train the Static Facial Expressions in the Wild (SFEW), to address the problem of small size of SFEW dataset.

S.Ouellet~\cite{11} used a CNN based architecture for realtime emotion detection. The author uses transfer learning to train the Cohn-Kanade~\cite{12} dataset on AlexNet. The author used the model to capture emotion of gamers while playing games. Ghimre et al.~\cite{15} used landmark points to generate a grid on the face. The localized regions were used for feature extraction using two methods: LBP and normalized central moments. The extracted features were used in SVM for classification. 

\textit{\textbf{Spatial transformer layer (STL)}}:
CNN is a very powerful model, invariant to some transformations like in-plane rotation and scaling. To obtain such invariance, CNN requires a huge amount of training data. To achieve such invariance in a computationally efficient manner, spatial transformer network~\cite{jaderberg2015spatial} is used as the input layer, called here as the spatial transformer layer. This allows spatial manipulation of the data within the network. This differentiable module, when combined with the CNN, infuses invariance to rotation, scaling, and translation, with less training data than that with normal CNN.

\textit{\textbf{Sobel and Laplacian operators}}: Interest points and edges give substantial information about the content of an image. The gradient of an image at any point gives the direction of the maximum change. The Laplacian of an image highlights the regions of rapid changes in the intensity. The gradient and Laplacian of an image $f(x,y)$, denoted by $\Delta f(x,y)$ and $\Delta^{2} f(x,y)$, can be approximated by applying Sobel~\cite{sobel19683x3} and Laplacian~\cite{haralick1992computer} operators on an image, We have taken the input images from the dataset and applied Sobel and Laplacian operators on them to obtain their first and second derivatives, respectively. They also detect the intensity discontinuities as contours.

\textit{\textbf{Global average pooling and DepthSep layers}}: The architecture selected by us uses elegant techniques such as global average pooling (GAP) and depthwise separable convolution (DepSep) layers. The GAP layer has multiple advantages over the dense layer: (i) it reduces over-fitting to a large extent; (ii) huge reduction in the number of parameters compared to dense layer; (iii) the spatial average of feature maps is fed directly to the soft-max layer. The latter enforces better correspondence between the feature maps and the categories.

\textit{\textbf{Depthwise separable convolution (DepSep)}} : The advantage of using depthwise separable convolution layer is that it greatly reduces the number of parameters compared to the convolution layer. At a particular layer, let the total number of filters be N, the depth of the feature maps be D, and the size of the filter (spatial extent) used be $S_{e}$. In such a case, the total number of parameters in normal convolution is $S_{e} \times S_{e} \times D \times N$. DepSep is a two-step process: (i) filters of size $S_{e} \times S_{e} \times 1$ are applied to each feature; therefore, the total number parameters at this step is $S_{e} \times S_{e} \times D$; (ii) then, N filters of size $ 1 \times 1 \times D$ are applied. So, the number of parameters required at this step are $D \times N$. Combining steps (i) and (ii), the total number of parameters in DepSep layer are $S_{e} \times S_{e} \times D + D \times N $. Hence, the reduction in the number of parameters compared to normal convolution at each layer, where convolution is replaced by DepSep convolution, is: $\frac{S_{e} \times S_{e} \times D + D \times N }{S_{e} \times S_{e} \times D \times N} = \frac{1}{N} + \frac{1}{S_{e}^2}$

\section{Datasets used for the study}
We have used the KDEF~\cite{KDEF1998} and FERplus~\cite{ferplus} datasets for our experiments. KDEF dataset contains a total of 4900 images (divided into the 7 classes of neutral, anger, disgust, fear, happiness, sadness, and surprise), with equal number of male and female expressions. The FERplus dataset contains nearly 35000 images divided into 8 classes, including contempt. The FERplus dataset improves upon the FER dataset by crowd-sourcing the tagging operation. Ten taggers were asked to choose one emotion per image, which resulted in a distribution of emotions for each image. The training set contains around 28000 images. The remaining are divided equally into validation and test sets. The original image size is 48$\times$48 pixels.

\section{Experiments and results}
\label{experiments}
\begin{table}
	\centering
	\caption{FER results of RTNN and its various modifications proposed by us (parallel networks) on the KDEF dataset.}
	\resizebox{0.45\textwidth}{!}
	{
	\begin{tabular}{|l |c|} \hline
	\bf Architecture Details  & \bf Accuracy \%  \\ \hline
	Original RTNN by Arriaga et al. \cite{codepaper} & 	\bf 83.16   \\ \hline
	STL + RTNN 	&	84.08   \\ \hline
	RTNN + Laplacian RTNN           	 		&84.39   \\  \hline
	STL with RTNN + Gradient RTNN	 		&85.10   \\ \hline
	STL with RTNN + Laplacian RTNN        	&85.51   \\ \hline
	STL with Original, Gradient and Laplacian RTNN  & \bf 88.16   \\ \hline
	\end{tabular}
	}
	\label{ourskdef}
\end{table}

\begin{table}[t!]
\centering
\caption{FER accuracies on FERPlus dataset and the number of parameters (in millions) of our models vs. VGG13 (models are trained 4 times and their maximum, minimum and average accuracies are reported)}
    
\resizebox{0.45\textwidth}{!}{
\begin{tabular}{|l|c|c|c|c|}
\hline
 	\textbf{Models} & \textbf{Avg} & \textbf{Min} & \textbf{Max} & Parameters\\
 	\hline
VGG13 (our implementation) 	& 83.56	& 82.99 &	84.08 & 8.75\\
\hline
VGG13 + Laplacian (input concatenated)	& 86.22 &	85.94 &	86.56 & 8.75 \\
\hline
VGG13 + Sobel (input concatenated)	& 86.42	& 86.08	& 86.55 & 8.75 \\
\hline
VGG13 (reported) & 83.85	& 83.15 &	84.89 & 8.75 \\
\hline
\end{tabular}
}

\label{vggoriginalwithoursresults}
    \vskip -0.1in
\end{table}

We have carried out three different experiments. The first two are on improving the recognition accuracies of existing methods. In the third part, we have developed a computationally efficient architecture that achieves state-of-the-art results. Our initial experiments are performed on KDEF dataset. We found that there is consistent improvement in terms of recognition accuracy as shown by Table~\ref{ourskdef}. \textit{To confirm and establish that these performance gains are achieved due to the use of our features, we have performed experiments on the state of the art techinques~\cite{codepaper} and ~\cite{ferplus}.} 

\begin{figure}[ht!]
    \centering
\includegraphics[width=0.36\textwidth,height=0.15\textheight]{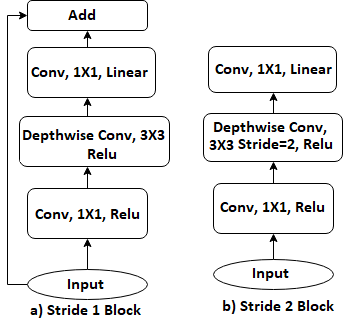}
\caption{Inverted bottleneck module used in MobileNetV2~\cite{mobilenetv2}.}
    \label{InvertedBottleneck}
\end{figure}

\begin{table}[]
 \caption{Details of architecture developed by us, with \emph{inverted bottleneck} \ref{InvertedBottleneck} as the core module. \emph{c}, \emph{s}, \emph{t} denote the number of output channels from each layer, stride and  expansion factor used in the bottleneck module, respectively.}
    \centering
\resizebox{0.45\textwidth}{!}{
\begin{tabular}{|c|c|c|c|c|c|}
\hline
Layer	& Parameters	& Input$(H\times W\times C)$ & c & s & t \\
 \hline
$conv2d$	& 480	& $ 64\times64\times1$  & 48 & 1 & --\\
$conv2d$	& 13856	& $64\times64\times48$  & 32 & 1 & --\\
$bottleneck$	& 14016 &	$64\times64\times32$ & 32 & 2 & 6\\
$bottleneck$	 & 12480 &	$32\times32\times32$ & 24 & 2 & 6\\
$bottleneck$	& 8208  &	$16\times16\times24$ & 24 & 1 & 6\\
$bottleneck$	& 9360  & $16\times16\times24$ & 32 & 2 & 6\\ 
$bottleneck$	& 14016 &	$8\times8\times32$ & 32 & 1 & 6\\ 
$bottleneck$	& 14016 &	$8\times8\times32$ & 32 & 1 & 6\\ 
$bottleneck$	& 20160 &	$8\times8\times32$ & 64 & 1 & 6\\
$bottleneck$	& 52608 &	$8\times8\times64$ & 64 & 1 & 6\\ 
$bottleneck$	& 52608 &	$8\times8\times64$ & 64 & 1 & 6\\
$bottleneck$	& 52608 &	$8\times8\times64$ & 64 & 1 & 6\\
$bottleneck$	& 77184 &	$8\times8\times64$ & 128 & 1 & 6\\
$bottleneck$	& 301824 & $8\times8\times128$  & 256 & 1 & 6\\
$avg\_pool$	& 0	& $8\times8\times256$ & 256 & -- & --\\
$conv2d$	& 2048	& $1\times1\times256$ & 8 & 1 & --\\
\hline
Total	&645472 & --	& -- & -- & --\\
\hline
\end{tabular}
}
\label{ourmodellesscomplexparameters}
\end{table}

\begin{table}[]
 \caption{FER performance of our models, with less complexity than VGG13, on FERPlus~\cite{ferplus} dataset. Models are trained 4 times with same hyper-parameter settings and their average, maximum and minimum accuracies are reported.}
    \centering
\resizebox{0.45\textwidth}{!}{
\begin{tabular}{|c|c|c|c|c|}
\hline
 	\textbf{Models} & \textbf{Avg} & \textbf{Min} & \textbf{Max} & Parameters \\
 	\hline
base model (given in Table III)	& 80.37	& 79.99 &	80.61 & 0.65 million \\
\hline
base + Laplacian (input concatenated)	& 82.83 &	82.38 &	83.33 & 0.65 million \\
\hline
base + Sobel (input concatenated)	& 83.3	& 82.71	& 83.99 & 0.65 million\\
\hline
base + Laplacian (parallel network)	& 82.71	& 81.96 &	83.15 & 0.65  million\\
\hline
base + Sobel (parallel network)	& 82.57	& 82.34	& 82.89  & 0.65 million \\ 
\hline
\end{tabular}
}
\vspace{-0.15 in}
\label{ourferparallel}
\end{table}

\textbf{1:} \textit{Real-time neural network (RTNN)} is the model proposed by Arriaga et al. \cite{codepaper}, trained on the KDEF~\cite{KDEF1998} dataset and validated. \textit{STL + RTNN} refers to the RTNN model trained with the use of an additional spatial transformer layer at the input. \textit{RTNN + Laplacian RTNN} refers to the architecture, where the input image and its Laplacian are fed in parallel. The outputs of these parallel sub-networks are combined and passed to a soft-max layer for classification. \textit{STL with RTNN + Gradient RTNN} is the case when the input image and its gradient are first fed to a STL, followed by the parallel subnetworks. The two parallel subnetworks are used for extracting more useful features in the beginning layer. Later, they are combined to obtain better accuracy. \textit{STL with RTNN + Laplacian RTNN } is the architecture, where the model is trained in parallel with the input image and its Laplacian. \textit{STL with Original, Gradient and Laplacian RTNN} is the architecture where the model is trained in parallel with three input image streams, namely, the original, the gradient, and the Laplacian images. It is to be noted that these input streams are first fed independently to a STL, before being fed to the subnetworks in parallel.

\textbf{2:} We have reimplemented the VGG13 network, used in ~\cite{ferplus}, in Tensorflow. We use the majority voting technique, as described in~\cite{ferplus}, for labelling each image. The only modification we have made to the original model is that we have used Adam optimizer \cite{adam} instead of momentum optimizer. In our setup, we get an average accuracy of 83.56\% instead of 83.85\% as reported in the original paper. Next we propose two experimental setups. First, we modify the input by taking Laplacian of original image and channel wise concatenating it with the original image. The resultant image is a 2 channel 64*64 input. In this setup, without modifying the learning rate, we get an average accuracy improvement of close to 3\% on an average, compared to our VGG13 implementation. In the second setup, we use Sobel operator instead of Laplacian. Using Sobel operator, we get gradients in 2 directions x and y. The resultant gradients are again concatenated to the original image channel wise to get 3 channel input. This setup again gives an improvement of close to 3\%. 

\textbf{3:} We propose our own architecture (details listed in Table~\ref{ourmodellesscomplexparameters}, having (1/13)-th the number of parameters compared to all the architectures reported in Table~\ref{vggoriginalwithoursresults}), developed using inverted bottleneck module (refer Fig.~\ref{InvertedBottleneck}) reported in~\cite{mobilenetv2}. The results with this base model and its variants are listed in Table IV. The \textit{base+Sobel} model performs comparable to the original VGG13 model listed in Table~\ref{vggoriginalwithoursresults}. 

\section{Conclusion}

We have proposed techniques to improve the performance of any FER system. We have performed many experiments on KDEF and FERplus datasets and achieved better recognition accuracy than the state-of-the-art techniques~\cite{codepaper} and~\cite{ferplus}. We have also developed a computationally efficient architecture that gives similar performance with one-tenth the number of parameters. We have shown that feeding the gradient and/or Laplacian of the image, in addition to the input image, gives better recognition accuracy. The advantages of our approaches are: (i) the dataset size increases by two or three times (depending on the Laplacian or/and gradient used together with the input), which is desirable in most deep learning tasks; (ii) the variability in the input image space increases (iii) DepSep, inverted bottleneck module and GAP layers help in reducing the computational complexity of the model. The proposed enhancements result in absolute performance improvements, as listed in Tables \ref{ourskdef}, \ref{vggoriginalwithoursresults}, and \ref{ourferparallel}, over those of the original models. We have shown that incorporating the gradient and Laplacian of the image dataset improves the accuracy of any deep learning based architecture for FER tasks. The approaches used in this work give consistent performance gain as validated by the Tables of results. 

One might argue that the gradient and Laplacian of the input image can very well be computed by the CNN. However, there are strong evidences for the need for appropriate representations in accomplishing certain vision and motor control tasks \cite{schomaker}. It is also clear from the results that at least the networks proposed by Arriaga et al.~\cite{codepaper} and VGG13~\cite{ferplus} are not able to compute these derived images internally. On the other hand, pre-computing these features and feeding them to the same network in parallel or in series, is clearly able to improve the performance of the network. Thus, our experiments show that there is a clear case for optimally combining appropriate feature extractors with learning neural networks, to obtain better performance for specific pattern recognition tasks.  


%


\ifCLASSOPTIONcaptionsoff
  \newpage
\fi



%

%

\end{document}